\documentclass[oneside]{article}

\usepackage{PRIMEarxiv}

\usepackage[utf8]{inputenc} 
\usepackage[T1]{fontenc}    
\usepackage{hyperref}       
\usepackage{url}            
\usepackage{booktabs}       
\usepackage{amsfonts}       
\usepackage{nicefrac}       
\usepackage{microtype}      
\usepackage{lipsum}
\usepackage{fancyhdr}       
\usepackage{graphicx}       
\graphicspath{{media/}}     
\usepackage{multirow}
\usepackage{bigstrut}

\usepackage{algorithm}
\usepackage{amsmath}
\usepackage{adjustbox}
\usepackage{multirow}
\usepackage{amssymb}
\usepackage{anyfontsize}
\usepackage{blindtext}
\usepackage{multirow}
\usepackage{amssymb}
\usepackage{pifont}
\usepackage[dvipsnames]{xcolor}
\usepackage[dvipsnames]{xcolor}
\usepackage[labelformat=simple]{subcaption}

\DeclareCaptionLabelFormat{subcaptionlabel}{\normalfont(\textbf{#2}\normalfont)}
\newcommand{\xmark}{\ding{55}}%
\usepackage{algpseudocode} 
\usepackage{booktabs} 

\title{
Transitivity-Preserving Graph Representation Learning for Bridging Local Connectivity and Role-based Similarity }

\author{
  Van Thuy Hoang and O-Joun Lee${}^{\dagger}$ \\
  Department of Artificial Intelligence, The Catholic University of Korea \\
  Bucheon-si, Gyeonggi-do 14662, Republic of Korea\\
  \texttt{\{hoangvanthuy90,ojlee\}@catholic.ac.kr} \\
  }

\begin{document}
\maketitle

\begin{abstract}

Graph representation learning (GRL) methods, such as graph neural networks and graph transformer models, have been successfully used to analyze graph-structured data, mainly focusing on node classification and link prediction tasks.
However, the existing studies mostly only consider local connectivity while ignoring long-range connectivity and the roles of nodes.
In this paper, we propose Unified Graph Transformer Networks (UGT) that effectively integrate local and global structural information into fixed-length vector representations.
First, UGT learns local structure by identifying the local substructures and aggregating features of the $k$-hop neighborhoods of each node.
Second, we construct virtual edges, bridging distant nodes with structural similarity to capture the long-range dependencies.
Third, UGT learns unified representations through self-attention, encoding structural distance and $p$-step transition probability between node pairs. 
Furthermore, we propose a self-supervised learning task that effectively learns transition probability to fuse local and global structural features, which could then be transferred to other downstream tasks. 
Experimental results on real-world benchmark datasets over various downstream tasks showed that UGT significantly outperformed baselines that consist of state-of-the-art models.
In addition, UGT reaches the expressive power of the third-order Weisfeiler-Lehman isomorphism test (3d-WL) in distinguishing non-isomorphic graph pairs.
The source code is available at \url{https://github.com/NSLab-CUK/Unified-Graph-Transformer}.

\let\thefootnote\relax
\footnotetext{${}^{\dagger}$ Correspondence: \texttt{ojlee@catholic.ac.kr}; Tel.: +82-2-2164-5516} 
\end{abstract}


\section{Introduction}\label{sect:intro}

Graph neural networks (GNNs) and graph transformers have effectively solved graph analysis tasks, such as node classification and link prediction \cite{DBLP:journals/corr/KipfW16, dwivedi2020generalization, s23084168}.
GNNs learn representations by iteratively aggregating features of neighbouring nodes with uniform or attentional weights \cite{DBLP:journals/corr/KipfW16,DBLP:journals/sensors/JeonCL22}.
Unlike GNNs, graph transformer models learn node representations using self-attention, encoding pair-wise connections between nodes, and have shown better performance than variants of GNNs \cite{dwivedi2020generalization}.

\begin{figure}[t]
\centering 
  \includegraphics[width=.4\linewidth]{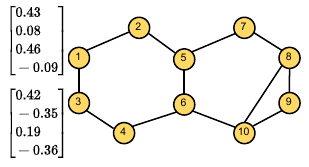}
  \caption{
 The Laplacian positional encoding cannot guarantee capturing similarity between surrounding structures of nodes.
 Two adjacent nodes, `1' and `3', have different positional encodings even though they are structurally similar.
  }
  \label{fig:example_laplac}
\end{figure}


While the existing models have been applied to various domains, GNNs and graph transformers still have three inherent limitations.
\textbf{{First}}, graph transformer models lack the power to capture similarity between local structures.
Some recent graph transformers (e.g., GT \cite{dwivedi2020generalization} and SAN \cite{DBLP:conf/nips/KreuzerBHLT21}) use graph structural features (e.g., Laplacian Eigenvectors) as positional encodings (PEs) to distinguish isomorphic substructures.
However, although the PEs can assign distinguishable representations to substructures of individual nodes, they do not always reflect structural similarity between the substructures.
PEs that account for substructure similarity are significant for the models to understand connectivity between neighbours, rather than simply aggregating their feature vectors. 
In Figure \ref{fig:example_laplac}, we assign the PEs to two nodes, `1' and `3', with Laplacian Eigenvectors.
Then, their vector representations are dissimilar, although they have similar local structures.
This impreciseness can be a serious obstacle to learning representations of graph structures and eventually affect the models' performance.

\textbf{{Second}}, GNN aggregators and self-attention lack the power to learn global structural features.
GNN variants, like GCN \cite{DBLP:journals/corr/KipfW16}, GAT  \cite{velivckovic2017graph}, and GATv2 \cite{DBLP:journals/corr/abs-2105-14491,Lee2020,lee2020learning}, compose node embeddings based only on features of neighbours. 
Similarly, graph transformer models learn node representations by encoding features of node pairs and substructures within $k$-hop distance using self-attention \cite{DBLP:conf/icml/ChenOB22,lee2020story,jeon2021learning}.
Although graph transformers have larger aggregation ranges in a layer than GNNs, neighbourhood aggregation is not powerful enough to reveal long-range dependencies or role-based similarities of nodes. 


\textbf{{Third}}, the existing models lack the ability to integrate local and global structural features into a unified representation, which could benefit various downstream tasks. 
For example, influential nodes in social networks will be centres of different communities, and we consider both `influence (global)' and `community (local)' to analyse the nodes. 
Some GNNs and graph transformers, such as LSPE \cite{DBLP:conf/iclr/DwivediL0BB22} and GPS \cite{DBLP:conf/nips/RampasekGDLWB22}, try to capture higher-order substructures surrounding nodes within $k$-hop distance and consider them in messages or node features.
However, we cannot increase the $k$-hop distance as the graph diameter increases. 
Furthermore, global structural features (e.g., long-range dependencies) correlate with node roles (i.e., substructures rooted in nodes). 
Similar roles of nodes do not always indicate their adjacency or similar node features but rather the opposite.
Therefore, the different views of the features are a significant barrier to integrating these features into a unified vector representation.

To overcome the limitations, we propose Unified Graph Transformer Networks (UGT), which can represent both local and global structural features of graphs with a unified vector representation. 
We first construct virtual edges connecting the distant nodes with role-based similarity. 
By sampling $k$-hop neighbourhoods with both virtual and actual edges, UGT can observe long-range dependencies of nodes along with their local connectivities. 
We propose structural identity, which is a substructure descriptor, and use it for identity mapping in transformer layers to preserve information about roles of nodes. 
Structural distances estimated based on the substructure descriptor are also used to compute attention scores to consider role-based similarity. 
Transition probabilities are used together to calculate the attention scores to consider proximity and role jointly.
Finally, to bridge the gap between local and global structures, UGT is pre-trained to preserve $p$-step transition probabilities, considering all the paths between two nodes and reflecting proximity and surrounding structure in multiple scales. 
Our contributions are as follows:
\begin{itemize}
    \item We propose Unified Graph Transformer Networks (UGT), which can learn local and global structural features and integrate them into a unified representation.
    \item Our novel sampling method constructs virtual edges between nodes with high role-based similarity and samples $k$-hop neighbourhoods to observe long-range dependencies and local connectivity together. 
    \item The pre-training task is proposed to bridge the conceptual gap between local and global structures by preserving transition probabilities between nodes, which reflect both connectivity and surrounding structures of the nodes from all the paths connecting them, at multiple scales. 
    \item We experimentally validated that UGT achieves comparable expressive power to the third-order Weisfeiler-Lehman isomorphism test (3d-WL) in distinguishing non-isomorphic graph pairs.
\end{itemize}

\section{Related Work}





This section discusses how the existing studies attempted to overcome the limitations discussed in the previous section. 
Several existing graph transformers attempt to reveal similarity of local structures between neighbouring nodes using PEs. 
SAN \cite{DBLP:conf/nips/KreuzerBHLT21} uses learnable PEs combined with the attention mechanism to learn the local structures. 
Graphformer \cite{DBLP:conf/nips/YingCLZKHSL21} adds node degree to node features and integrates edge features using shortest-path distance attention bias.
GPS \cite{DBLP:conf/nips/RampasekGDLWB22} uses random walk positional encoding (RWPE) to find similar substructures for each target node.
SAT \cite{DBLP:conf/icml/ChenOB22} extracts $k$-hop subgraph representations for each target node and uses GNNs to update the target node representations. 
The existing models try to consider similarity of local structures using substructures or paths surrounding nodes without preserving distances between target and context nodes, although close neighbours usually have higher significance than distant ones \cite{DBLP:journals/mis/LeeHK21,DBLP:journals/joi/LeeJJ21}. 
The main difference between the existing studies and UGT is that we describe and consider substructures while preserving the distances using the structural identity and distance. 
By describing the distances, the structural identity goes beyond local structure descriptors to cover global structures, including the roles of nodes.


Several existing studies attempted to capture global structural features.
WRGAT \cite{DBLP:conf/kdd/SureshBNLM21} aims to break the limitations of GNNs by using multiple relations between nodes.
GeoGCNs \cite{DBLP:conf/iclr/PeiWCLY20} employ multiple embedding methods, such as Poincare and Struc2vec, and choose an optimal latent space that can preserve the global structural features. 
While the GNN variants show the capability to capture global information, they mostly overlook substructure similarity.
Several graph transformers, such as SAT \cite{DBLP:conf/icml/ChenOB22} and Graphformer \cite{DBLP:conf/nips/YingCLZKHSL21}, attempt to capture higher-order structures within $k$-hop distance \cite{nguyen2023connector}.
However, the higher-order structures are extracted within $k$-hop distance and omit long-range dependencies since the $k$-hop distance is difficult to be extended to graph diameters. 
Otherwise, UGT constructs virtual edges to connect distant nodes with substructure similarity. 
By observing occurrences of substructures all over the graph with virtual edges and co-occurrences of substructures within $k$-hop distance with actual edges, UGT can learn both local and global structures. 
Then, the transition probabilities are the medium for fusing the two features.


\section{Unified Graph Transformer Networks}

\begin{figure*}[t]
\centering 
  \includegraphics[width=1\linewidth]{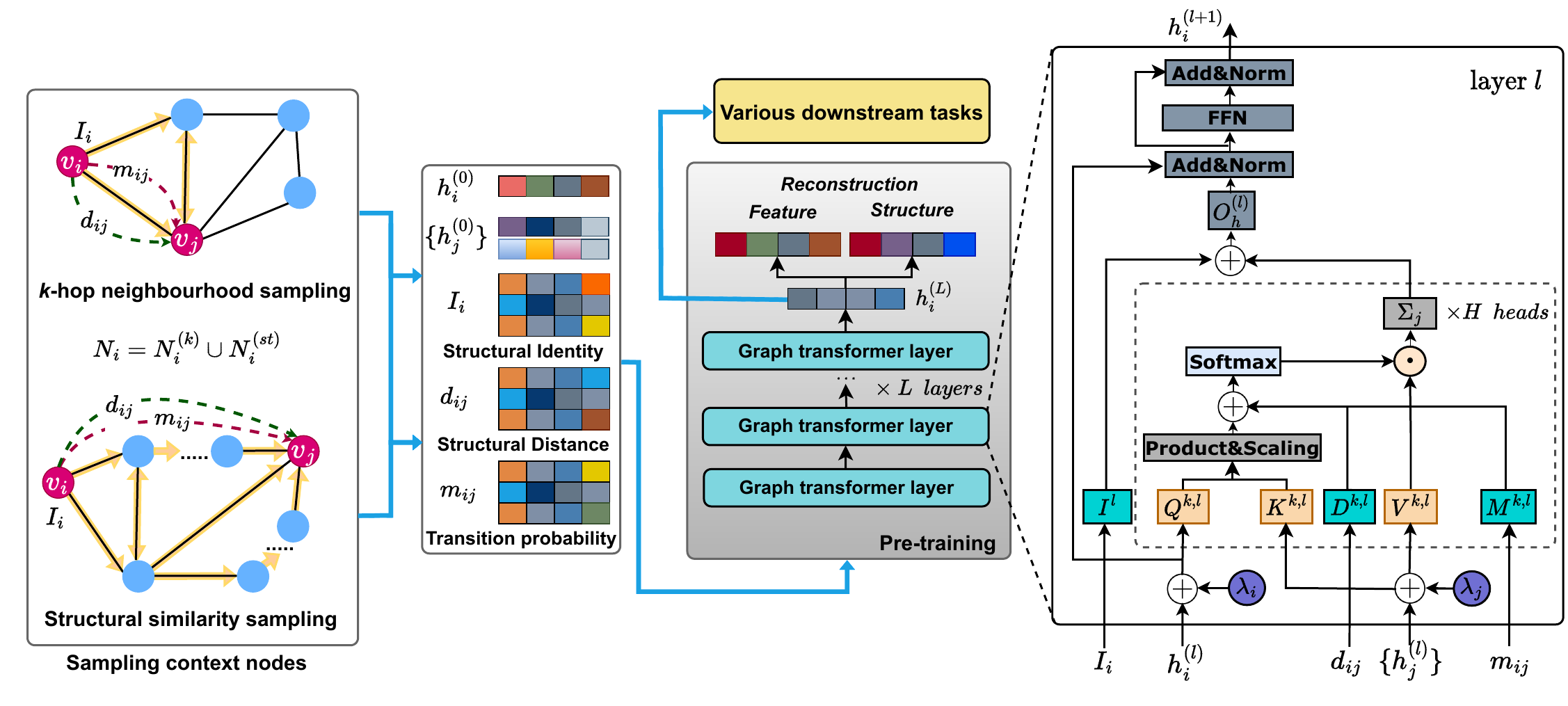}
  \caption{The overall architecture of UGT.
  UGT is composed of main blocks, including sampling context nodes, building modules $I$, $d$, and $m$, and pre-training blocks. 
  The learned representations then could be used for various downstream tasks.
  }  
  \label{fig:model}
\end{figure*}

This section introduces a novel graph transformer model, UGT, which can learn local and global structural features and integrate them into a unified vector representation.
We first introduce how to sample context nodes and then describe the design of UGT in detail.
Finally, we introduce self-supervised learning and fine-tuning tasks.
The overall architecture of UGT is described in Figure \ref{fig:model}.

\subsection{Sampling Context Nodes } 

\begin{figure}[t]
\centering 
  \includegraphics[width=.5\linewidth]{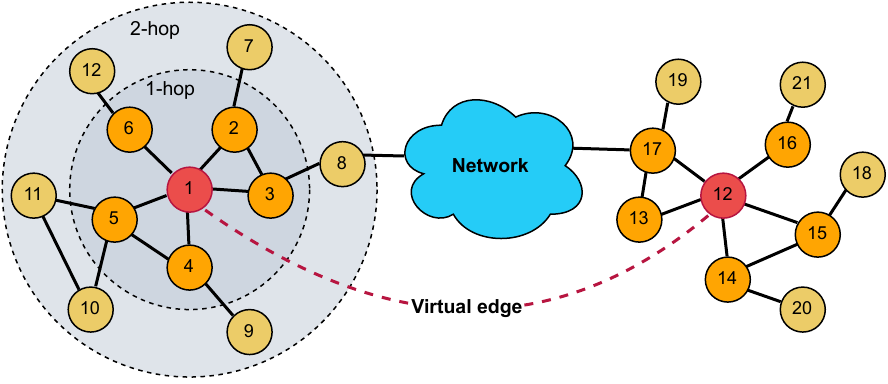}
  \caption{An example of UGT sampling strategy. 
  We sample context nodes within $k$-hop neighbourhoods and virtual connections between distant nodes with structural similarity. }
  \label{fig:sample}
\end{figure}

Given a graph $G = (V, E)$ with node set $V$ and edge set $E$, we aim to sample the context node $v_j$ of each target node $v_{i} \in V$.
The context node $v_{j}$ could be sampled if the distance from $v_{i}$ to $v_{j}$ within $k$-hop, or both two nodes $v_{i}$ and $v_{j}$ are structurally similar.
Formally, for each target node $v_{i}$, the set of neighbours of $v_{i}$ can be defined as:
\begin{eqnarray}
    {N}_{{i}}&=& N^{(k)}_{{i}} \cup N^{(st)}_{i},
\end{eqnarray}
where $N^{(k)}_{{i}}$ indicates the set of neighbours of $v_{i}$ within $k$-hop distance, and $N^{(st)}_{i}$ refers to the set of nodes that are structurally similar to $v_{i}$.
Figure \ref{fig:sample} presents a simplified graph that contains a target node `1' and its context.
For global sampling, we construct virtual edges between any two nodes if they are structurally similar.
For example, nodes `1' and `12' are structurally similar with the same degree of five and connected to two triangles.
Note that we only consider similarity based on the graph structure without using the node feature.
Let $I_{k}({v_{i}})$ denotes the set of an ordered degree sequence of $v_{i}$ in $1$ to $k$-hop.
Note that $I_{0}({v_{i}})$ is the degree of $v_{i}$.
By comparing the ordered degree sequences between $v_{i}$ and $v_{j}$, we can impose a score to measure structural distance.
Formally, the structural distance and score between any two nodes $v_{i}$ and $v_{j}$ in a graph could be defined as:

\begin{align}
& s_k = {{{f}_{k}}\left( I_{k}({v_{i}}),I_{k}({v_{j}}) \right)}, \\
& S \left(v_{i}, v_{j} \right ) = {\exp \left ({-\sqrt{s_k}} \right)},
\end{align}
where $f_k (\cdot , \cdot)$ is the structural distance between two sequences.
Similar to Struc2vec \cite{ribeiro2017struc2vec}, we use dynamic time warping (DTW) to measure the distance between two ordered degree sequences.
In most real-world networks, however, the degree distribution is highly asymmetric and shows a long tail distribution.
To boost the connections between nodes with high degrees, we introduce another score:
\begin{align}
& S \left(v_{i}, v_{j} \right ) = {\exp \left ({-\sqrt{s_k}}\right )}+{\exp \left ( {-\frac{1}{^{\sqrt{\left( {{d}_{v_i}} + {{d}_{v_j}} \right)}}}}\right )},
\end{align}
where $d_{v_i}$ presents the degree of $v_{i}$.
Accordingly, two nodes with a high score are structurally similar, and if they have high degrees together, they are closer in the latent space.

\subsection{Learning Unified Representations of Graphs}


\subsubsection{Input representation}

Given a graph $G$, the node feature $x_{i} \in R^{d_{0} \times 1}$ of node $v_{i}$ is mapped via a linear projection to $d$-dimensional hidden feature $\hat{x}_{i}^{0}$, as:
\begin{eqnarray}
    \hat{x}_{i}^{0}&=&{{W}_{0}}{{x}_{i}}+{{b}_{0}},
\end{eqnarray}
where ${W}_{0} \in R^{d \times d_{0} }$ and ${b}_{0} \in R^{d}$ are the parameters of the linear projection, $d_0$ refers to the original feature of $v_{i}$.
Since we aim to make UGT can learn the global positional information context for each node in the whole graph, we add a linearly transformed positional embedding $\lambda_{i}$ of dim $k$ to node features, as:
\begin{eqnarray}
    \hat \lambda _{i}&=&{{W}_{1}}{{\lambda }_{i}}+{{b}_{1}}, \\
    {h}_{i}^{0}&=& \hat{x}_{i}^{0}   + \hat  \lambda _{i},
\end{eqnarray}
where ${W}_{1} \in R^{d \times k }$ and ${b}_{1} \in R^{d}$.
Note that the positional embeddings are pre-calculated and added only for the first time, and we use Eigenvectors of the graph Laplacian matrix.
Furthermore, we randomly flip $\lambda$ during training to allow UGT to capture sign invariance.


\subsubsection{Global structural self-attention } 

We now propose a self-attention bias to encode local and global structural information. 
Since we aim to capture context nodes within a $k$-hop distance and distant nodes, the self-attention needs to understand the distance between node pairs in terms of $k$-hop distance and the structural distance.
Formally, we define a self-attention between $v_{i}$ and $v_{j}$ for head $k$ at layer $l$ as:
\begin{align}
  & \alpha _{ij}^{k,l}=\frac{ \left({{Q}^{k,l}}h_{i}^{l}\right) \cdot \left({{K}^{k,l}}h_{j}^{l}\right)}{\sqrt{d_k}}+ D_{ij}^{k,l} + M_{ij}^{k,l},
\end{align}
where ${Q}^{k,l}$, ${K}^{k,l}$ $\in R^{d_{k} \times d}$, $k = 1$ to $H$ refers to the number of attention heads, $h_{i}$ and $h_{j}$ are the features of node $v_{i}$ and $v_{j}$, respectively.
$D_{ij}^{k,l}$ and $M_{ij}^{k,l}$ $\in R^{d_k \times d}$ are linear transformed structural distance $d_{ij}$ and transition probability $m_{ij}$ between two nodes $v_i$ and $v_j$, respectively.
We now introduce strategies to construct $d$ and $m$.

We define the structural identity of each target node based on hierarchical role-based information, including minimum, maximum, mean, and standard deviation degrees up to $k$-hop distance context of each node $v_i$.
Formally, the structural identity of node $v_{i}$ and the structural distance could be defined as:
\begin{align}
   &{I}_{i}= \{ \  {{d}_{v_i}}, T_1 , T_2, \cdots , T_k\}, \\
  & {{d}_{ij}}= f \left( \left[ {{\left\| I_{i}^{(q)}-I_{j}^{(q)} \right\|}_{2}} \right]^{-1} \right), 
\end{align}
where $d_{v_i}$ denotes the degree of node $v_{i}$, 
$ T_k =\{ {{\min}_{k}},{\max}_{k},{{\mu }_{k}},{{\delta }_{k}}\}$ denotes the set of minimum, maximum, mean, and standard deviation values of the node degrees at $k$-th hop distance,
$I_{i}^{(q)}$ refers to the $q$-th element of $I_{i}$, 
and $f$ refers to a linearly transformed structural distance. 

While structural distance can benefit the model by finding distant nodes with structural similarity, it cannot capture paths between the two nodes.
To fill this gap, we present the transition probability-based distance $m$, which is the awareness of connectivity between any two nodes from a probabilistic perspective.
In this way, UGT could capture both the local and global structural information, leading to the power to capture any path between nodes.
We define the transition probability distance $m$ from $v_{i}$ to $v_{j}$ as:
\begin{align}
 & m_{ij} = f\left( {A}_{ij}^{1},{A}_{ij}^{2},\cdots ,{A}_{ij}^{p} \right), 
\end{align}
where ${A}_{ij}^{p}$ refers to the transition probability from $v_{i}$ to $v_{j}$ at $p$-step,
and $f$ refers to a linearly transformed transition probability.

\subsubsection{Graph transformer layers}

The outputs of self-attention are then concatenated into one vector representation followed by a linear transformation.
The node features of $v_i$ are updated at layer $l$ as:
\begin{equation}
\label{eq:12}
\hat h_{i}^{l+1}=O_{h}^{l}\underset{k=1}{\overset{H}{\mathop{\mathbin\Vert}}}\,\left( \sum\limits_{v_j\in N({{v}_{i}})}{\tilde{\alpha }_{ij}^{k,l}{{V}^{k,l}}h_{j}^{l}} \right),
\end{equation}
where ${\tilde{\alpha }_{ij}^{k,l}} = \text{softmax}_{j} ({{\alpha }_{ij}^{k,l}}  ) $, ${Q}^{k,l}$, ${K}^{k,l}$,${V}^{k,l}$ $\in R^{d_{k} \times d}$, ${O}^{l}_{h} \in  R^{d \times d}$, and $\mathbin\Vert$ refers to concatenation.
The outputs then are passed to feed-forward networks (FFN) along with residual connections and layer normalization:
\begin{align}
  & \hat{h}_{i}^{l+1}=\text{LN}\left( h_{i}^{l}+\hat{h}_{i}^{l+1} \right), \\ 
 & \hat{\hat{h}}_{i}^{l+1}=W_{2}^{l}\text{ReLU}\left( W_{1}^{l}\hat{h}_{i}^{l+1} \right), \\ 
 & h_{i}^{l+1}=\text{LN}\left( \hat{h}_{i}^{l+1}+\hat{\hat{h}}_{i}^{l+1} \right),  
\end{align}
where $W_{1}^{l} \in R^{2d \times d}$ and $W_{2}^{l} \in R^{d \times 2d}$ are learnable parameters, and LN refers to layer normalization.


Since the structural identity of nodes could sufficiently extract information around nodes based on degree information, it could benefit to distinguish non-isomorphic substructures.
Therefore, we add a linearly transformed structural identity to each transformer layer, as given by Equation (\ref{eq:12}):
\begin{align} 
&\hat h_{i}^{l+1}=f\left( I_{i}^{l} \right)+ \hat h_{i}^{l+1}.
\end{align}
We believe that with a number of adequate transformer layers, UGT could map different substructures into different representations.


\subsection{Self-supervised Learning Tasks}



We now present self-supervised learning tasks that could train UGT on pretext tasks to extract graph structure without using any label information.
Our objective is to learn representations that could capture both local and global 
structures between any nodes as long as they are structurally similar.
As mentioned earlier, we aim to preserve relational structure information in $p$-step transition probability, combining local and global structures.
The learned representations could then be used for solving different downstream tasks.

Given a connectivity between a target node $v_i$ and a context node $v_j$, we aim to maximize the transition probability of paths connecting $v_i$ and $v_j$ against the probability of other pairs not from the graph. 
Similar to Grarep and Glove, we employ noise contrastive estimation, introduced by \cite{GutmannH12}.
The loss function for the transition probability matrix of $v_i$ at $p$-step could be defined as:
\begin{align}
  & {{L}^{p}_{1}}({{v}_{i}})=\left( \sum\limits_{{{v}_{j}}}{{P_{k}}\left( {{v}_{j}}|{{v}_{i}} \right)\log \sigma \left( {{z}_{i}}{{z}_{j}} \right)} \right) + \lambda \mathbb{E} , 
\end{align}
where  $\mathbb{E} =  {{E}_{{{v}_{k}}\sim {P_{k}}(V)}} \left [ \log \sigma \left( -{{z}_{i}}{{z}_{k}} \right) \right]$, $v_{k}$ refers to nodes obtained from negative sampling.
We then assume the noise follows a uniform distribution, and this lead to a new transition matrix on a log scale that represents precisely the global relation information between any two nodes at step $p$:
\begin{align}
 & A'^{(p)}=\log \left( \frac{A_{i,j}^{(p)}}{\sum\nolimits_{t}{A_{t,j}^{(p)}}} \right)-\log \left( \frac{\left| N_{s} \right|}{\left| V \right|}  \right),
\end{align}
where $A'^{(p)}$ is the log scale probability matrix at step $p$ that we aim to preserve, and $N_{s}$ refers to the set of negative samples.
Finally, the loss function for structure preservation at step $p$ can be computed as:
\begin{align}
 L^{p}_{1} =  \left\| {A^{'(p)} - Z^{(p)}})\right\|_{F}^{2}, 
\end{align}
where ${{{Z}}^{(p)}}$ is the score matrix that could be built by computing the cosine similarity between vector embeddings.

Since node feature information could benefit several downstream tasks, UGT should also learn a reconstruction of node feature information.
We define the node raw feature reconstruction-based loss term as:
\begin{align}
 {L_{2}}=\frac{1}{\left| V \right|}\sum\nolimits_{{{v}_{i}}\in V}{{{\left\| {{x}_{i}}-{{{\hat{x}}}_{i}} \right\|}_{2}}}, 
\end{align}
where $x_{i}$ refers to the raw feature of node $v_{i}$, and $\hat{x_{i}} = FFN (z_{i})$.
We train the model in multi-task learning with the two losses, and the losses are linearly combined as:
\begin{align}
 L = \alpha \sum\limits_{p}{L^{p}_{1}}  + \beta L_2, 
\end{align}
where $\alpha$ and $\beta$ are hyper-parameters.

\subsection{Fine-tuning Tasks}

To apply UGT to downstream tasks, the learned representations are then passed directly to solve downstream tasks.
In this study, we present three downstream tasks, including node clustering, node classification, and graph-level classification.
For the node clustering task, we used modularity as the loss function as with \cite{DBLP:journals/jmlr/TsitsulinPPM23}.
For the node classification, a representation of node $v_i$ from the transformer layers was passed to fully-connected (FC) layers to get the classification output $y_i$ as:
\begin{align}
y_i  =  W_1 \text{ ReLU} \left( W_2 h^{l}_{i} \right), 
\end{align}
where $W_1  \in  R^{d \times C}$ and $W_2 \in R^ {d \times d}$ are weight matrixes, and $C$ refers to the number of classes.
In the graph-level classification, we employed average pooling to obtain graph features and likewise used FC layers to obtain the prediction output.

\section{Experiments}

We conducted experiments to evaluate UGT versus GNN variants and graph transformer models for three tasks: node clustering, node classification, and graph-level classification.
We also analyzed the power of our model by assessing UGT on isomorphism testing.

    


\subsection{Experimental Settings}

\subsubsection{Datasets}
For the node-level tasks, we used eleven publicly available datasets, which are grouped into three different domains, including Air-traffic networks (e.g., Brazil, Europe, and USA) \cite{ribeiro2017struc2vec}, Webpage networks (e.g., Chameleon, Squirrel, Actor, Cornell, Texas, and Wisconsin) \cite{DBLP:conf/iclr/PeiWCLY20}, and Citation networks (e.g., Cora and Citeseer) \cite{sen2008collective}.
We used four publicly available datasets for the graph classification task, including Enzymes, Proteins, NCI1, and NCI9 from TUDataset \cite{Morris_2020}. 
Furthermore, we used Graph8c and nine Strongly Regular Graphs datasets (SRGs), which contain 1d-WL and 3d-WL equivalent graph pairs, respectively, for isomorphism testing \cite{DBLP:conf/icml/BalcilarHGVAH21}.

\subsubsection{Baselines} 
The GNN variants included GCN~\cite{DBLP:journals/corr/KipfW16}, GCNII~\cite{pmlr-v119-chen20v}, GIN~\cite{DBLP:conf/iclr/XuHLJ19}, GAT~\cite{velivckovic2017graph}, GATv2~\cite{DBLP:journals/corr/abs-2105-14491}, SAGE~\cite{DBLP:conf/nips/HamiltonYL17}, Geom-S~\cite{DBLP:conf/iclr/PeiWCLY20}, WRGAT~\cite{DBLP:conf/kdd/SureshBNLM21}, and DeeperGCN~\cite{DBLP:journals/corr/abs-2006-07739}.
Furthermore, we compared UGT against recent transformers, including GT~\cite{dwivedi2020generalization}, SAN~\cite{DBLP:conf/nips/KreuzerBHLT21}, SAT~\cite{DBLP:conf/icml/ChenOB22}, GPS~\cite{DBLP:conf/nips/RampasekGDLWB22}, ANS-GT~\cite{DBLP:conf/nips/Zhang0HL22}, and Graphormer~\cite{DBLP:conf/nips/YingCLZKHSL21}.
For isomorphism testing, we compare UGT against recent powerful models, i.e.,  ChebNet \cite{DBLP:conf/nips/DefferrardBV16}, and GNNML3 \cite{DBLP:conf/icml/BalcilarHGVAH21}.



\subsubsection{Implementation Details}
We conducted each experiment ten times by randomly sampling  training, validation, and testing sets of size 80\%, 10\%, and 10\%, respectively.
The results written in the tables were measured with means and standard deviation on the testing set over the ten cases.
The experiments were done in two servers with four NVIDIA RTX A5000 GPUs (24GB RAM/GPU).
The hyper-parameters were tuned on the validation sets for each task and dataset.
For fair comparisons with the baselines, we conducted a search for the number of layers and the hidden dimension size with ranges of \{2, 4, 8\} and \{32, 64, 128\} with every model, respectively. 


\begin{table*}[t]
\centering
\fontsize{13 pt}{16 pt}\selectfont
\begin{adjustbox}{width=1\textwidth}
\begin{tabular}{l cc cc cc cc cc cc cc cc cc cc cc}
    \toprule
\multirow{1}{*}{} 
        & \multicolumn{2}{c}{Brazil} & \multicolumn{2}{c}{Europe} & \multicolumn{2}{c}{USA} & \multicolumn{2}{c}{Chameleon} & \multicolumn{2}{c}{Squirrel} & \multicolumn{2}{c}{Film} & \multicolumn{2}{c}{Cornell} & \multicolumn{2}{c}{Texas}  & \multicolumn{2}{c}{Wisconsin} & \multicolumn{2}{c}{Cora}& \multicolumn{2}{c}{Citeseer}   \\
        \cmidrule(lr){2-3} \cmidrule(lr){4-5} 
        \cmidrule(lr){6-7} \cmidrule(lr){8-9}
        \cmidrule(lr){10-11} \cmidrule(lr){12-13} 
        \cmidrule(lr){14-15} \cmidrule(lr){16-17}  \cmidrule(lr){18-19} 
        \cmidrule(lr){20-21} \cmidrule(lr){22-23}  

         &C$\downarrow$& Q$\uparrow$
         &C$\downarrow$& Q$\uparrow$ 
         &C$\downarrow$& Q$\uparrow$ 
         &C$\downarrow$& Q$\uparrow$ 
         &C$\downarrow$& Q$\uparrow$
         &C$\downarrow$& Q$\uparrow$ 
         &C$\downarrow$& Q$\uparrow$
         &C$\downarrow$& Q$\uparrow$
         &C$\downarrow$& Q$\uparrow$
         &C$\downarrow$& Q$\uparrow$
         &C$\downarrow$& Q$\uparrow$\\
    \midrule \hline
GCN &0.71 & -0.01 & 0.71 & -0.07 & 0.47 & 0.16 &\textcolor{orange}{\textbf{ 0.46 }}& \textbf{ 0.25} & \textbf{ 0.27} &\textcolor{orange}{\textbf{ 0.73}}& 0.59 & \textbf{0.11} & \textbf{0.75} & \textbf{0.0} & \textcolor{orange}{\textbf{0.61}} & \textcolor{orange}{\textbf{0.16}} & \textcolor{orange}{\textbf{0.59}} &\textcolor{orange}{\textbf{ 0.18}} & \textbf{0.16} & \textbf{ 0.68} & \textbf{0.17 }& \textbf{0.63}   \\

GCNII & 0.74 & -0.02 & 0.76 & -0.03 & 0.59 & 0.11 & 0.56 & 0.2 & 0.72 & 0.27 & 0.66 & 0.06 & 0.78 & -0.05 & 0.8 & -0.03 & \textbf{0.64} & \textbf{ 0.09} & 0.42 & 0.33 & 0.32 & 0.37 \\

GIN& 0.73 & -0.03 &\textcolor{red}{\textbf{ 0.01}} & \textcolor{orange}{\textbf{0.01}} & \textcolor{red}{\textbf{0.08}} & 0.1 & \textbf{0.47} & 0.13 & 0.75 & 0.27 & 0.59 & 0.04 & 0.81 & -0.1 & 0.9 & -0.13 & 0.73 & -0.02 & 0.2 & 0.63 & 0.24 & 0.56  \\

GAT& 0.68 & -0.01 & 0.74 & -0.05 & 0.69 & 0.03 & 0.66 & 0.09 & 0.77 & 0.25 & 0.71 & 0.02 & 0.85 & -0.1 & 0.88 & -0.18 & 0.82 & -0.07 & 0.17 & 0.66 & 0.25 & 0.55 \\

GATv2 & 0.54 & -0.01 & 0.61 & -0.06 &{\textbf{ 0.22}} & \textbf{0.17} & 0.72 & 0.06 & 0.77 & 0.2 & 0.7 & 0.06 & 0.8 & -0.08 & 0.88 & -0.13 & 0.72 & -0.01 & 0.17 & 0.66 & 0.21 & 0.59 \\

SAGE& 0.55 & -0.02 & 0.76 & -0.06 & 0.68 & 0.04 & 0.77 & 0.03 & 0.75 & 0.25 & 0.72 & 0.0 & 0.9 & -0.16 & 0.94 & -0.15 & 0.83 & -0.07 & 0.18 & 0.65 & 0.23 & 0.57 \\

WRGAT & 0.74 & -0.01 & 0.78 & -0.05 & 0.71 & 0.03 & 0.64 & 0.09 & 0.69 & 0.34 & 0.67 & -0.0 & 0.86 & -0.15 & 0.92 & -0.17 & 0.83 & -0.12 & 0.23 & 0.6 & 0.28 & 0.53 \\

WRGCN & 0.72 & -0.0 & 0.79 & -0.05 & 0.7 & 0.03 & 0.64 & 0.1 & 0.68 & 0.34 & 0.67 & 0.0 & 0.88 & -0.16 & 0.92 & -0.17 & 0.79 & -0.07 & 0.24 & 0.59 & 0.27 & 0.53 \\

DeeperGCN &\textcolor{orange}{\textbf{0.46}} & 0.0 & \textbf{0.55} & -0.06 & 0.5 & 0.1 & 0.75 & 0.04 & 0.78 & 0.26 & 0.75 & 0.03 & 0.8 & -0.07 & 0.88 & -0.15 & 0.75 & -0.05 & 0.18 & 0.65 & 0.23 & 0.57 \\\hline

GT &  0.61 & -0.01 & 0.6 & -0.05 & 0.31 & 0.12 & 0.78 & 0.02 & 0.78 & 0.22 & 0.8 & -0.01 & 0.87 & -0.15 & 0.92 & -0.19 & 0.81 & -0.11 & 0.19 & 0.64 & 0.26 & 0.54\\

SAN & 0.6 & -0.04 & 0.66 & -0.06 & 0.42 & 0.15 & 0.71 & 0.01 & 0.58 & \textbf{ 0.42} &\textcolor{orange}{\textbf{ 0.42}} & 0.01 & 0.89 & -0.15 & 0.88 & -0.12 & 0.79 & -0.04 & 0.23 & 0.61 & 0.25 & 0.55\\

GPS & 0.7 &\textcolor{orange}{\textbf{ 0.03}} & 0.7 & \textbf{-0.01} & 0.64 & -0.0 & 0.7 & 0.08 & 0.81 & 0.2 & 0.65 & 0.0 & 0.8 & -0.09 & 0.73 & -0.05 & 0.78 & -0.05 & 0.38 & 0.46 & 0.5 & 0.29 \\

SAT & \textcolor{red}{\textbf{0.43 }}& \textbf{0.01} & 0.75 & -0.06 & 0.72 & -0.0 & 0.8 & -0.08 & 0.64 & 0.35 & 0.73 & -0.0 & 0.85 & -0.07 &\textbf{ 0.7} & \textbf{ 0.02} & 0.81 & -0.04 & 0.58 & 0.01 & 0.39 & 0.12 \\\hline


UGT(K-mean) & 0.68   & 0.0  &0.78 & -0.05  &\textcolor{orange}{\textbf{0.13}} & \textcolor{orange}{\textbf{0.22}} & \textcolor{red}{\textbf{0.11}} &\textcolor{red}{\textbf{ 0.66 }}&\textcolor{orange}{\textbf{0.24}} & \textcolor{red}{\textbf{0.74}}& \textbf{0.44} & \textcolor{orange}{\textbf{0.32}} & \textcolor{orange}{\textbf{0.52}} & \textcolor{orange}{\textbf{0.1}} & 0.8 & -0.1 & 0.67 & 0.03 & \textcolor{orange}{\textbf{0.10}} & \textcolor{orange}{\textbf{0.71}} & \textcolor{orange}{\textbf{0.15}} & \textcolor{orange}{\textbf{0.66}} \\

UGT & \textbf{0.51}  & \textcolor{red}{\textbf{0.22}}  & \textcolor{orange}{\textbf{0.51}}  & \textcolor{red}{\textbf{ 0.20}} & 0.34 & \textcolor{red}{\textbf{0.30}} & \textcolor{red}{\textbf{0.11}} & \textcolor{orange}{\textbf{0.64}} & \textcolor{red}{\textbf{0.21}} & 0.39 & \textcolor{red}{\textbf{0.28}} & \textcolor{red}{\textbf{0.50}}& \textcolor{red}{\textbf{ 0.28 }}& \textcolor{red}{\textbf{0.47}} & \textcolor{red}{\textbf{0.33}} & \textcolor{red}{\textbf{0.46}} & \textcolor{red}{\textbf{0.27}}  & \textcolor{red}{\textbf{0.52}} & \textcolor{red}{\textbf{0.09}} & \textcolor{red}{\textbf{0.76}} & \textcolor{red}{\textbf{0.04}} & \textcolor{red}{\textbf{0.78}} \\

\bottomrule
\end{tabular}
\end{adjustbox}
\caption{The performance on node clustering task in terms of conductance (C) and modularity (Q) measurements.
The top three are emphasized by \textcolor{red}{\textbf{first}}, \textcolor{orange}{\textbf{second}}, and \textbf{third}.   }
\label{tab:node_clustering}
\end{table*}

\subsection{Performance Analysis}
\subsubsection{Evaluation on node clustering}
We first conducted an experiment on eleven benchmark datasets for the node clustering task.
For the baselines, we trained the models in a supervised manner with a classification task. Then, we used learned representations as inputs for the K-means clustering algorithm.
For our UGT model, we conducted clustering using both K-means and an end-to-end manner.
Table \ref{tab:node_clustering} shows the performance of node clustering in terms of conductance (C) and modularity (Q) measurements \cite{DBLP:journals/kais/YangL15}. 
\textbf{(1)}
UGT outperformed all baselines that ignore either local or global structures, e.g., GCN, WRGAT, and GT.
Remarkably, the performance of our learned representations combined with K-mean also showed significant performance over the benchmark datasets.
We suppose that as UGT could learn local and global structure and graph density, the representations could estimate the dense relations between nodes in communities.
\textbf{(2)} Most of the recent graph transformers, such as SAN, SAT, Graphormer, and GPS, did not capture the density and connectivity information, only concerning higher-order neighbourhoods.
This indicates that learning higher-order substructures could not help the model handle local relation density and graph partitioning.


\begin{table*}[t]
\fontsize{22 pt}{28 pt}\selectfont
\begin{adjustbox}{width=1\textwidth}

  \begin{tabular}{l ccc cccccccc}
    \toprule
     &Brazil & Europe & USA&Chameleon & Squirrel & Film & Cornell & Texas & Wisconsin & Cora & Citeseer  \\\hline \hline
    GCN & 41.54$\pm$9.73 &  37.94$\pm$2.99& 50.42$\pm$5.07   & 66.25 $\pm$1.77 & \textbf{50.92 $\pm$1.02}&28.15$\pm$1.37 & 55.55$\pm$9.29 & 48.89$\pm$6.47 & 58.40$\pm$6.49 & 87.33$\pm$1.71 & 72.05$\pm$1.22 \\
    
    GCNII& 24.35$\pm$3.62 &  22.78$\pm$5.37 &28.29$\pm$4.19&60.07$\pm$2.44& 28.30$\pm$1.63& 26.03$\pm$0.77&  50.92$\pm$6.54& 75.00$\pm$2.26& 60.67$\pm$2.49& 84.22$\pm$0.74& 70.22$\pm$2.25 \\
    
     GIN &  50.76$\pm$7.84 &  42.56$\pm$9.67 & 50.75$\pm$9.77  & 66.87$\pm$2.72  & 40.53$\pm$1.16 & 23.21$\pm$1.13 & 36.66$\pm$7.53 &34.44$\pm$13.78 & 44.00$\pm$12.64 & 77.25$\pm$3.35 & 64.09$\pm$1.95 \\
    
    GAT & 58.46$\pm$7.84&  54.35$\pm$3.40 & 46.38$\pm$11.41   & \textbf{67.84$\pm$0.8} & \textcolor{orange}{\textbf{64.76$\pm$0.72}}& 26.21$\pm$1.44 & 57.77$\pm$6.66 & 68.88$\pm$9.02 & 60.00$\pm$11.02 & 84.29$\pm$2.02 & 73.43$\pm$1.21 \\
    
    GATv2 &{ 69.23$\pm$8.42} & 57.94$\pm$4.75 & 61.84$\pm$ 3.38  & 62.20$\pm$2.11 & 50.80$\pm$3.01 & 25.73$\pm$2.11 & 56.66$\pm$8.88 &  63.33$\pm$9.68 & 55.20$\pm$4.66 & 85.77$\pm$1.27& 72.95$\pm$1.87 \\
    
    SAGE & 66.15$\pm$7.84& \textbf{60.16$\pm$4.59} & 60.00$\pm$6.60 & \textcolor{orange}{\textbf{67.92$\pm$3.83}} & 47.42$\pm$1.07 &32.57$\pm$0.91 &\textcolor{orange}{\textbf{77.77$\pm$6.08}}& 82.22$\pm$10.18 &81.60$\pm$6.49 &\textbf{87.77$\pm$1.49} & \textbf{74.69$\pm$1.89} \\
    
    Geom-S & - & - &-  & 58.94$\pm$0.54  & 36.55$\pm$0.14 &30.46$\pm$0.32 & 54.05$\pm$0.00 & 64.86$\pm$0.00&58.82$\pm$0.00 & 83.40$\pm$0.10 & 73.50$\pm$0.07 \\
    
    WRGAT & 55.38$\pm$3.07 &  51.28$\pm$3.62 &  56.97$\pm$3.54 & 51.80$\pm$2.03 & 32.96$\pm$1.44 & 35.78$\pm$1.84 &\textbf{74.44$\pm$15.55} & 75.55$\pm$6.66 &\textcolor{orange}{\textbf{84.80$\pm$5.30}}& 74.59$\pm$0.50 & 73.01$\pm$1.71 \\
    
    
    DeeperGCN & \textbf{69.23$\pm$4.86}&  59.49$\pm$6.76 & 61.17$\pm$4.99  & 57.53$\pm$2.82 &34.96$\pm$1.04&31.68$\pm$1.50 & 63.33$\pm$10.30 &76.67$\pm$8.16& 72.80$\pm$8.54 & 86.59$\pm$0.85 & 73.61$\pm$0.62 \\\hline
    
    GT & 63.07$\pm$11.30 &\textcolor{orange}{\textbf{ 62.56 $\pm$9.53}} & 64.37 $\pm$ 2.67  & 65.55$\pm$2.83 & 49.50$\pm$ 1.59 & 34.55$\pm$ 1.90 & 70.37$\pm$ 5.24 &  \textbf{84.44$\pm$ 7.37} & 82.40$\pm$5.98 & 86.17$\pm$ 1.96 & 71.58$\pm$ 2.38 \\
    
     SAN & 61.53$\pm$10.87 &\textcolor{red}{\textbf{ 63.24$\pm$5.26}} & 35.29$\pm$3.43 & 
    64.02$\pm$2.60&
    46.28$\pm$2.09 &
    32.14$\pm$0.27& 
    \textcolor{red}{\textbf{79.62$\pm$2.61}}& 
    \textcolor{orange}{\textbf{85.18$\pm$9.44}}& 
    82.66$\pm$1.88& 
    84.81$\pm$1.98& 
    73.99$\pm$2.69 \\
    
     GPS & 52.31$\pm$8.97 & 46.00$\pm$5.15 & 42.52$\pm$3.85  & 42.54$\pm$3.87 & 34.42$\pm$2.14 &35.37$\pm$2.20 &45.05$\pm$7.75 &30.63$\pm$1.27 &62.00$\pm$9.09 & 62.12$\pm$1.00 & 51.18$\pm$3.35 \\
    
   SAT &66.15 $\pm $6.15 & 57.43$\pm$6.19 & \textcolor{orange}{\textbf{65.04$\pm$4.06}}& 49.69$\pm$3.81 &40.08$\pm$0.76& 31.61$\pm$1.37 & 41.67$\pm$6.33 & 34.44$\pm$4.15& 57.60$\pm$9.32& 79.63$\pm$2.69\ & 63.98$\pm$2.41 \\
    
    ANS-GT& 46.80$\pm$5.30 & 31.00$\pm$3.89& 46.20$\pm$3.31  & 54.60$\pm$1.02 &35.80$\pm$1.17 & \textcolor{red}{\textbf{39.80$\pm$0.75}} &59.80$\pm$2.71&63.80$\pm$5.04&\textcolor{red}{\textbf{85.20$\pm$2.71}}&\textcolor{orange}{\textbf{88.00$\pm$0.63}}&\textcolor{orange}{\textbf{75.00$\pm$0.89}} \\
    
    Graphormer &42.00$\pm$3.2 &29.80$\pm$1.17&45.80$\pm$3.66&53.80$\pm$1.17  &34.60$\pm$0.80& \textcolor{orange}{\textbf{39.20$\pm$0.75}}  & 66.20$\pm$1.47& 74.00$\pm$4.86 &
    \textbf{84.00$\pm$1.26} & 86.40$\pm$0.49& 74.60$\pm$0.80\\\hline
    
    UGT w/o PT &\textcolor{orange}{\textbf{76.92$\pm${4.71}}} 
    & 44.94 $\pm$4.86 
    & {\textbf{ 64.71$\pm$3.19}}  
    & 60.33 $\pm$3.91 
    & 50.43  $\pm$0.81 
    & 23.63  $\pm$1.03
    &  52.78 $\pm$6.51 
    &   63.89  $\pm$3.93 
    & 64.0 $\pm$7.21
    & 74.63 $\pm$2.93  
    &57.41 $\pm$1.25  \\
    
     UGT  
     & \textcolor{red}{\textbf{80.00 $\pm$5.23}}
     & 56.92 $\pm$6.36
     &\textcolor{red}{\textbf{ 66.22$\pm$4.55}} 
     & \textcolor{red}{\textbf{ 69.78 $\pm$3.21}} 
     &\textcolor{red}{\textbf{  66.96 $\pm$2.49}} 
     & \textbf{36.84$\pm$0.62} 
     & 70.0 $\pm$4.44
     &  \textcolor{red}{\textbf{86.67 $\pm$8.31}} 
     & 81.6  $\pm$8.24
     & \textcolor{red}{\textbf{ 88.74$\pm$0.6 }} 
     & \textcolor{red}{\textbf{ 76.08$\pm$2.5}} \\
    
    \hline
\end{tabular}
\end{adjustbox}
\caption{The performance on node classification task (accuracy). 
The top three are emphasized by \textcolor{red}{\textbf{first}}, \textcolor{orange}{\textbf{second}}, and \textbf{third}. }
\label{tab:node_classification}
\end{table*}

\subsubsection{Evaluation on node classification}

We show the results of the baselines and UGT on the node classification task in Table \ref{tab:node_classification}.
\textbf{(1)} Our UGT model with pre-training outperformed baseline models that overlook the long-term dependencies in most datasets, i.e., GCN, GAT, GATv2, and GT.
We assume that the virtual edges allowed UGT to capture the long-term dependencies between structurally similar nodes, which have analogous roles and are significant for analyzing heterophily graphs.
\textbf{(2)}  UGT showed significant improvements in homophily graphs, i.e., Cora and Citeseer, in which nodes with the same labels tend to be adjacent.
This indicates that UGT could not only capture local connectivity but also distinguish similar substructures (roles).

\begin{table}[t]
\centering
\fontsize{10 pt}{13 pt}\selectfont
\begin{adjustbox}{width = 0.5 \textwidth}

  \begin{tabular}{l  cccc}
    \toprule
    & Enzymes & Proteins & NCI1 & NCI109  \\\hline \hline
    GCN& 18.24$\pm$2.05 &  59.23$\pm$0.62 & 68.24$\pm$2.38&  67.09 $ \pm$ 3.43 \\
    SAGE & 21.46$\pm$4.32 &  62.79$\pm$1.38 &  64.36$\pm$2.82 &  64.47$\pm$2.41 \\
    GCNII &31.46$\pm$5.31& 62.53$\pm$2.41&  63.27$\pm$1.38& 68.12$\pm$0.54 \\
    GIN & 33.64$\pm$3.52 & 64.14 $ \pm$ 2.05&  66.72$\pm$5.32 &68.44$\pm$1.89 \\
    GATv2 &  25.17$\pm$4.42&  66.85$\pm$ 2.43&  61.58$\pm$1.43&  64.51$\pm$2.36 \\
    DeeperGCN  &25.36$\pm$4.79 &  61.24$\pm$3.59 & 55.32$\pm$3.28&  55.1$\pm$2.18 \\\hline
    
    GT & 41.67$\pm$6.67 & \textcolor{orange}{\textbf{77.25$\pm$3.83 }}& \textbf{69.77$\pm$1.4}&  \textbf{69.66$\pm$0.06}\\
    SAN & 22.50$\pm$0.83 &\textbf{68.47$\pm$0.9 }& 59.31$\pm$3.47  & 57.3$\pm$8.20\\
    SAT& {\textbf{50.85$\pm$3.66  }}&62.91$\pm$0.76 &54.99$\pm$0.26 &  56.04$\pm$0.06\\
    GPS &\textcolor{orange}{\textbf{ 62.71$\pm$8.64}} & 53.75$\pm$6.20 &  \textcolor{red}{\textbf{79.44$\pm$0.65  }}& \textcolor{red}{\textbf{76.27$\pm$0.95}}\\\hline
    
    UGT  & \textcolor{red}{\textbf{67.22$\pm$3.92}} &\textcolor{red}{\textbf{80.12 $\pm$0.32}} & \textcolor{orange}{\textbf{77.55 $\pm$0.16 }}& \textcolor{orange}{\textbf{75.45$\pm$1.26}}\\\hline  
    \newline
\end{tabular}
\newline
\end{adjustbox}
\caption{The performance on graph classification task in terms of accuracy. 
  The top three are emphasized by \textcolor{red}{\textbf{first}}, \textcolor{orange}{\textbf{second}}, and \textbf{third}.   }
  \label{tab:graph_classification}
\end{table}



\subsubsection{Evaluation on graph-level classification}

Table \ref{tab:graph_classification} shows the performance of the models on graph-level classification.
For fair comparisons, we did not use the pre-training in the graph-level tasks, since graphs in the benchmarks have small scales.
UGT exhibited competitive results compared to GNN variants and graph transformers, showing the effectiveness in capturing the global structure information in the whole graphs.
This indicates that UGT could capture the connectivity in each individual substructure and then map unique substructures to different representations. 


\begin{table*}[t]
\centering
\caption{
The number of graph pairs that are undistinguished in Graph8c and nine 
strongly regular graphs (SRGs).
Since graphs in each benchmark dataset are non-isomorphic, an ideal graph representation model should not find any similar pairs.
The top two are emphasized by \textcolor{red}{\textbf{first}} and \textcolor{orange}{\textbf{second}}.
}

  \begin{adjustbox}{width=1\textwidth}
  \fontsize{10 pt}{13 pt}\selectfont
  \begin{tabular}{l  c  cc cccc ccc}
    \toprule
    & GRAPH8C & SR16622 &   SR251256 & SR261034 & SR281264 & SR291467 & SR351668 & SR351899 & SR361446 &  SR401224\\    \hline \hline

    $\#$ Graphs & 11,117 &2 &15 & 10 & 4 & 41 & 3854 & 227 & 180 & 28  \\
    
    $\#$ Comparisions &61.8M & 1 & 105 & 45 & 6 & 820 & 7,424,731 & 25,651 & 16,110 & 378  \\
    \cmidrule(lr){2-2} \cmidrule(lr){3-11} 
    
    Types &\multicolumn{1}{c}{1-d WL}  & \multicolumn{9}{c}{3-d WL} \\

     \hline
     
    GCN& 4,196 & 1  & 105  &  45  &6   &  820 & 7,424,731  & 25,651  & 16,110 &378  \\
    GA & 1,827 & 1  & 105  &  45  &6   &  820 & 7,424,731  & 25,651  & 16,110  &378  \\
     GIN& 559 & 1  & 105  &  45  &6   &  820 & 7,424,731  & 25,651  & 16,110  &378  \\
    ChebNet  &  \textcolor{orange}{\textbf{44}} & 1  & 105  &  45  &6   &  820 & 7,424,731  & 25,651  & 16,110  &378  \\
    PPGN  &  \textcolor{red}{\textbf{0}} & 1  & 105  &  45 &6   &  820 & 7,424,731  & 25,651 & 16,110  &378  \\
    GNNML3  &   \textcolor{red}{\textbf{0}} & 1  & 105  &  45  &6   &  820 & 7,424,731  & 25,651  & 16,110  &378  \\
    
    \hline

    GT &\textcolor{red}{\textbf{0}}& \textcolor{red}{\textbf{0}}  & \textcolor{red}{\textbf{0}}    &  \textcolor{red}{\textbf{0}}  &\textcolor{red}{\textbf{0}}   & \textcolor{red}{\textbf{ 0}} & \textcolor{orange}{\textbf{19}}  & \textcolor{red}{\textbf{0}}  & \textcolor{red}{\textbf{0}}  & \textcolor{orange}{\textbf{2}}  \\
    GT w/o $\lambda$ & 6,157 & 1  & 105  &  45  &6   &  820 & 7,424,731  & 25,651  & 16,110  &378  \\
 
    SAN  & 5,819 & 1  & 105  &  45  &6   &  820 & 7,424,731  & 25,651  & 16,110  &378  \\

    SAT& 286,360 &  \textcolor{red}{\textbf{0}} & \textcolor{orange}{\textbf{ 91}} & \textcolor{orange}{\textbf{36}} & \textcolor{orange}{\textbf{3 }}&\textcolor{orange}{\textbf{742}} &  {\textbf{ 6,973,038}} & \textcolor{orange}{{23,899}} & \textcolor{orange}{\textbf{15,066 }} & {\textbf{351}} \\
    
    GPS&   \textcolor{red}{\textbf{0}} & 1  & 105  &  45  &6   &  820 & 7,424,731  & 25,651  & 16,110  &378 \\
    GPS w/o RWPE &  1,517 & 1  & 105  &  45  &6   &  820 & 7,424,731  & 25,651  & 16,110  &378 \\\hline
    
    UGT w/o $\lambda$ & 355 & 1  & 105  &  45  &6   &  820 & 7,424,731  & 25,651  & 16,110  &378  \\
    
    UGT w/o I & 36,857.0 & 1  & 105  &  45  &6   &  820 & 7,424,731  & 25,651  & 16,110  &378  \\
    
    UGT       & \textcolor{red}{\textbf{0}} & \textcolor{red}{\textbf{0}}  & \textcolor{red}{\textbf{0}}   &  \textcolor{red}{\textbf{0}}  &\textcolor{red}{\textbf{0}}   &  \textcolor{red}{\textbf{0}}& \textcolor{red}{\textbf{0}}  & \textcolor{red}{\textbf{0}}  & \textcolor{red}{\textbf{0}} &\textcolor{red}{\textbf{0}} \\\hline
\end{tabular}
\end{adjustbox}
  \label{tab:isomorphic_testing}
\end{table*}

\subsubsection{Evaluation on the power of model}

\begin{figure}[t]
\centering 
\begin{subfigure}{0.8\textwidth}
\centering 
  \vspace{-7pt} 
  \includegraphics[width=.9\linewidth]{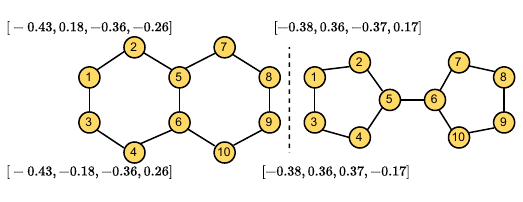}
  \vspace{-7pt} 
  \caption{Distinguishing non-isomorphic graphs with $\lambda$.}
  \label{fig:Lap_RWPE_a}
\end{subfigure}

\begin{subfigure}{0.8\textwidth}
\centering 
  \includegraphics[width=.9\linewidth]{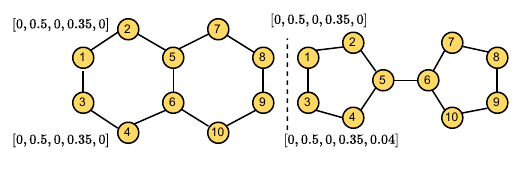}
  \vspace{-14pt}
  \caption{Distinguishing non-isomorphic graphs with RWPE.}
  \label{fig:Lap_RWPE_b}
\end{subfigure}\hfil 
\caption{Distinguishing a Decalin graph (Left) and Bi-cyclopentyl graph (Right).
}
\label{fig:Lap_RWPE}
\end{figure}

We conducted experiments to evaluate UGT on isomorphism testing for 1d-WL and 3d-WL benchmark datasets.
Table \ref{tab:isomorphic_testing} shows the performance on Graph8c and nine SRGs datasets.
\textbf{(1)} UGT could perfectly distinguish non-isomorphic graph pairs in Graph8c and SRGs.
Most of the GNN variants could also distinguish graphs in Graph8c.
However, as GNNs' capabilities are limited to the 1d-WL test, they failed to distinguish the graph pairs in SRGs.
\textbf{(2)} 
The structural identity was effective to distinguish 1d-WL graph pairs in Graph8c, which is more powerful than $\epsilon$ of GIN.
Since UGT determines substructures rooted in each node through structural identity ($I$), the learned representations thus could capture the surrounding substructures of each node.
\textbf{(3)}
To investigate the reasons why UGT is more powerful than baselines, we examined differences between Laplacian Eigenvectors and RWPE.
Figure \ref{fig:Lap_RWPE_a} presents how Laplacian Eigenvectors ($\lambda$) can distinguish a decalin and bi-cyclopentyl graph pair by assigning initial PEs to nodes `1' and `3'.
At the first iteration, the input representation of UGT can be represented as ${h}_{i}^{0} = \hat{x}_{i}^{0}+{{W}_{1}}{{\lambda }_{i}}+{{b}_{1}}$ and assigns different representations from the first iteration.
Thus, the output embeddings with structural identity ($h_{i}^{l+1} = FFN\left( I_{i}^{l} \right)+O_{h}^{l} (\cdot)$) ensure distinguishability to nodes. 
In contrast, RWPE needed a random walk with a length of five to distinguish the non-isomorphic graph pairs in Figure \ref{fig:Lap_RWPE_b}.
We suppose that in the SRGs, nodes are connected densely, which causes RWPE to fail to capture the substructures and connectivity in such graphs.

\subsection{Ablation studies}

To verify different modules in our model contribute to the overall performance, we conduct ablation studies on three benchmark datasets, including Cora, Brazil, and Cornell.
Table \ref{tab:ablation_1} shows the results of contributions of the five models: virtual edges, I, $\lambda$, d, and m.
We have the following observations:
\textbf{1) } Constructing virtual edges two components I and $\lambda$ could bring better performance in terms of accuracy for node classification tasks. 
We argue that as the Cora dataset is homophily graph, our model could understand the local and global structural information, thanks to the I and $\lambda$ components.
Furthermore, UGT can also capture the long-term dependencies in Brazil and Cornell as they are heterophily graphs.
This indicates that our model could effectively combine the virtual edges and two components.
\textbf{2) } Without virtual edges, our model can show good performance on clustering tasks.
We argue that constructing virtual edges may make UGT a bit confusing the connectivity in graphs since it can affect actual edges.
\textbf{4),} 
Considering the structural distance and transition probability could contribute to the highest performance of the model in terms of accuracy for the Cora dataset.
This is because the model could learn the local structural similarity and long-term dependencies even if we enable the virtual edges.
For understanding local connectivity, the transition probability (m) is beneficial compared to structural distance (d).
We argue that the transition probability could distinguish different connectivity between pair nodes in graphs, eventually helping the model to capture the density and the graph partitioning.

\begin{table*}[t]
\small
\centering

\begin{adjustbox}{width=1\textwidth}
\fontsize{15 pt}{19 pt}\selectfont

 \begin{tabular}{c  cc   ccc  ccc  ccc | cc ccc  ccc  ccc }
    \toprule
   \multirow{2}{*}{Virtual Edges}& 
   \multirow{2}{*}{I} &
   \multirow{2}{*}{$\lambda$} &

   \multicolumn{3}{c}{Cora} &
   \multicolumn{3}{c}{Brazil} &
   \multicolumn{3}{c | }{Cornell} &

    \multirow{2}{*}{D} &
   \multirow{2}{*}{M} &
   
    \multicolumn{3}{c}{Cora} &
   \multicolumn{3}{c}{Brazil} &
   \multicolumn{3}{c}{Cornell} \\

   \cmidrule(lr){4-6}  \cmidrule(lr){7-9} \cmidrule(lr){10-12}
   \cmidrule(lr){15-17}  \cmidrule(lr){18-20} \cmidrule(lr){21-23}

   \multicolumn{1}{l}{} & & & 
   acc &Q &C & acc &Q &C& acc &Q &C & & & acc &Q &C & acc &Q &C& acc &Q &C\\\hline \hline
   
    \multirow{4}{*}{ $\checkmark$ }
    
    & - &- & 86.87 & 0.71 & 0.14 & 76.92 & 0.06 & 0.49  & 50.0 &  0.02 &  0.62  
    & - &- & 85.00 & 0.70 & 0.15 & 76.92 &0.03 &0.30 & 50.0 &0.45 &  0.31 \\
    
     &$\checkmark$  & - &  87.06 & 0.73  & 0.12 & 76.92 & 0.09 & 0.52&\textbf{66.67} & 0.21 & 0.43  
     & $\checkmark$  & - 
     & 86.87 & 0.68 & 0.18 & \textbf{84.62} & \textbf{ 0.07} & 0.51 & 55.56 & 0.09 & 0.42\\
     
    &- &  $\checkmark$ &  86.50 & 0.72 & 0.13 & 69.23 & \textbf{-0.01} &\textbf{ 0.59} & 44.44 & 0.02 & 0.18 &  
    - &  $\checkmark$ & 
    86.69 & 0.74  & 0.11 & 84.62 & 0.07 & 0.46  & \textbf{61.11} & \textbf{0.43} & \textbf{0.34}\\
    
    &  $\checkmark$ &  $\checkmark$ & \textbf{87.80 }& \textbf{0.70} & \textbf{0.15} & 84.62 & 0.03 & 0.20 & 55.56 & 0.24 &  0.29  &
     $\checkmark$ &  $\checkmark$ & 
     \textbf{88.35} & 0.72 & 0.12 &76.92 & 0.07 & 0.37 &50.0 & 0.11 & 0.40    \\

    \hline
    \multirow{4}{*}{ \xmark } &
    - & - &  86.32 & 0.70 & 0.15 &  69.23 & -0.02 & 0.46 & 61.11 & 0.01 &  0.44  
    & - & - & 
    87.61  & 0.73  &0.12 & 76.92 & 0.02 &  0.13 & 55.56 & 0.55 & 0.23\\
    
     & $\checkmark$  & -&  87.24 &  0.72 & 0.13 & 84.62 & 0.07 & 0.37 & \textbf{66.67} & 0.19 &  0.29 &
     $\checkmark$  & -&
     86.50  & 0.73  &0.12 & 84.62  & 0.03 &0.20 & 50.0 &  0.17 &0.45\\
     
     & - &$\checkmark$  &  87.43 & 0.75 & 0.10 & 76.92 & 0.00 & 0.65 & 50.0 & \textbf{0.49} & \textbf{0.26}   & 
     - &$\checkmark$  &
     86.32 &   0.73  &0.12 & 76.92 & 0 &\textbf{ 0} & 50.0 & -0.16 & 0.66 \\
     
     & $\checkmark$  & $\checkmark$  & \textbf{85.40 }&\textbf{ 0.72} & \textbf{0.13} & 84.62 & 0.06 & 0.52 &  55.56 & 0.0 & 0.0  &
     $\checkmark$  & $\checkmark$  & 
     87.25 & \textbf{0.74} & \textbf{ 0.11} & \textbf{76.92} & \textbf{0.03} & \textbf{0.16} & 55.56 & 0.44 &0.25       \\
     \hline

\end{tabular}
\end{adjustbox}
\caption{Study the effect of virtual edges, I, $\lambda$, D, and M while keeping the other components in our model fixed.
The different consequences are marked and highlighted by colours.
}
\label{tab:ablation_1}
\end{table*}

\subsection{Sensitivity Analysis}

\begin{figure}[t]
    \centering 

\begin{subfigure}{0.8\textwidth}
  \includegraphics[width=\linewidth]{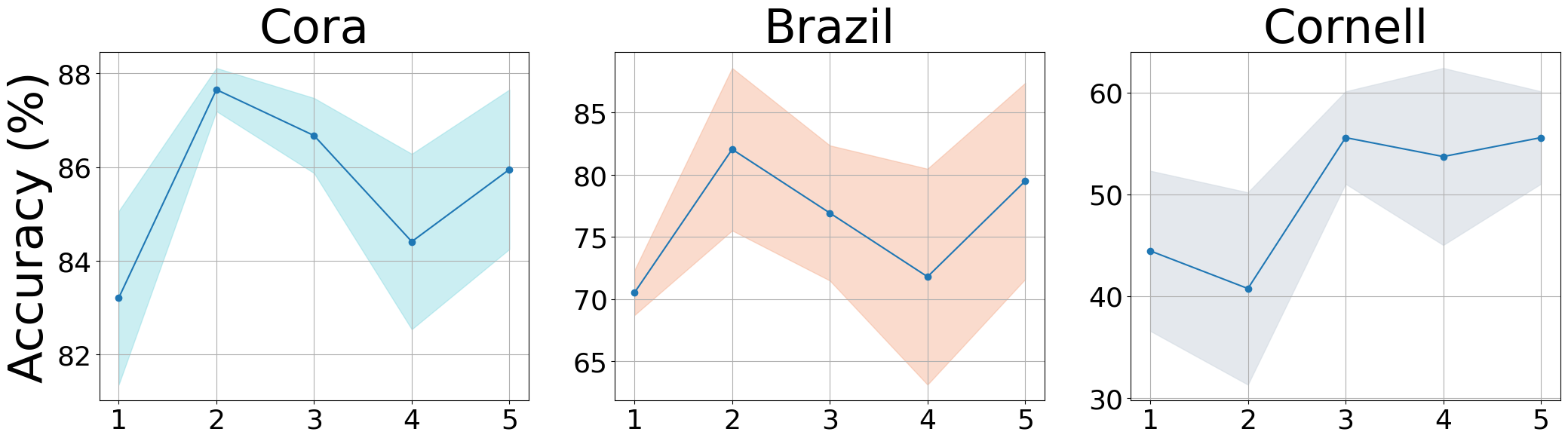}
  \centering \text{$k$-hop distance.}
  \label{fig:1}
\end{subfigure}\hfil 



\medskip

\begin{subfigure}{0.8\textwidth}
  \includegraphics[width=\linewidth]{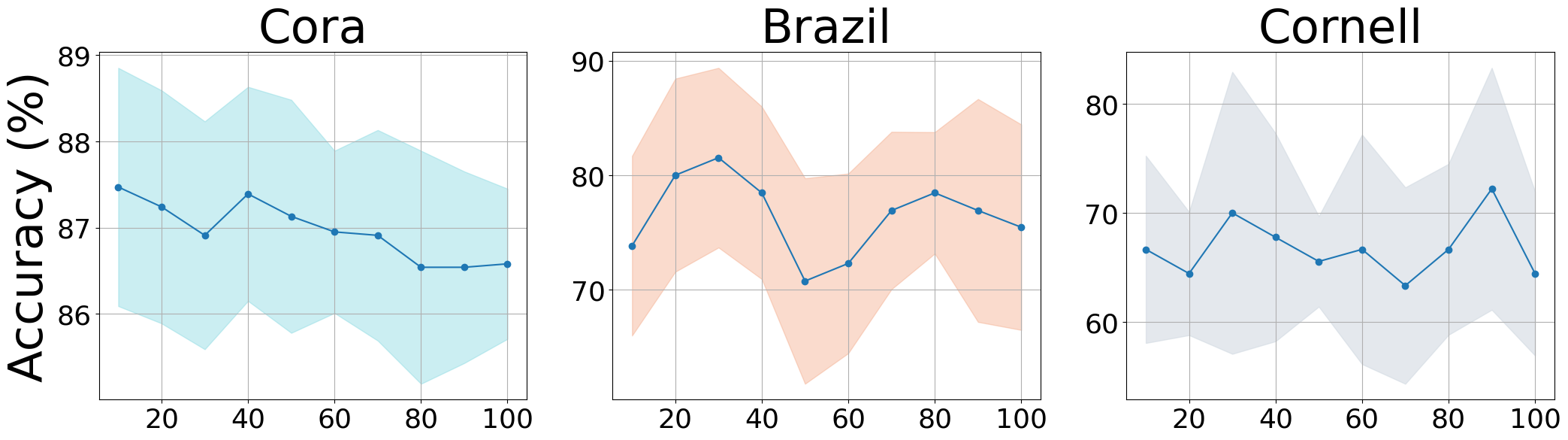}
  \centering \text{Virtual edges over actual edges.}
  \label{fig:4}
\end{subfigure}\hfil 

  \vspace{-7pt} 
\caption{The performance of node classification task according to the sampling range of context nodes up to $k$-hop distance (Above) and the percentage of virtual edges over actual edges (Below).}
\label{fig:Sensitivity}
\end{figure}

We performed sensitivity analyses on the range of $k$-hop neighbourhoods and the ratio of virtual edges to real edges, as shown in Figure \ref{fig:Sensitivity}.
\textbf{(1)} As $k$ (i.e., the range of neighbourhood sampling) increased, the performance of node classification showed increasing tendencies across the three datasets.
However, the trends were different for homophily and heterophily graphs.
For homophily graphs (e.g., Cora), increasing $k$ up to the 3-hop range affected model performance, but not beyond.
In contrast, for heterophily and sparse graphs, increasing the number of neighbourhoods provided more benefit.
\textbf{(2)} Adding virtual edges could benefit more than increasing the number of $k$-hop neighbourhoods for the Cora dataset.
We suppose that virtual edges could assist UGT in finding more distant nodes with structural similarity, and long-range dependencies could be effective for analyzing both homophily and heterophily graphs.

\section{Conclusion and future work}
\label{sec:conclusion}

This study proposes a novel graph transformer model, UGT, to learn local and global structural features and integrate them into a unified representation.
UGT captures long-range dependencies between nodes with similar roles using structural similarity-based sampling, discovers local connectivity using $k$-hop neighbourhoods and structural identity, and unifies them by learning transition probabilities between nodes that imply both aspects. 
Experimental results on various downstream tasks and datasets show that UGT outperforms or is comparable to state-of-the-art GNNs and graph transformers. 
Since our sampling technique uses the structural similarity between nodes in the whole graph, it can cause computational complexity in large graphs.
Our further research aims to reduce the computational complexity of role-based sampling by applying graph coarsening.


\section*{Acknowledgments}
This work was supported 
in part by the National Research Foundation of Korea (NRF) grant funded by the Korea government (MSIT) (No. 2022R1F1A1065516 and No. 2022K1A3A1A79089461) (O.-J.L.) 
and
in part by the Research Fund, 2022 of The Catholic University of Korea (M-2023-B0002-00088) (O.-J.L.). 

\bibliographystyle{unsrt}  
\bibliography{references}

\end{document}